\documentclass[pdflatex,sn-nature]{sn-jnl}

\usepackage{graphicx}%
\usepackage{multirow}%
\usepackage{amsmath,amssymb,amsfonts}%
\usepackage{amsthm}%
\usepackage{mathrsfs}%
\usepackage[title]{appendix}%
\usepackage{xcolor}%
\usepackage{textcomp}%
\usepackage{manyfoot}%
\usepackage{booktabs}%
\usepackage{algorithm}%
\usepackage{algorithmicx}%
\usepackage{algpseudocode}%
\usepackage{listings}%

\theoremstyle{thmstyleone}%
\theoremstyle{thmstyletwo}%

\theoremstyle{thmstylethree}%

\unnumbered

\begin{document}

\title{Vision-Language Model-Guided Deep Unrolling Enables Personalized, Fast MRI}


\author[1,3]{\fnm{Fangmao} \sur{Ju}}\email{fangmao@stu.xjtu.edu.cn}
\equalcont{These authors contributed equally to this work.}

\author[1,3]{\fnm{Yuzhu} \sur{He}}\email{yuzhu.he@stu.xjtu.edu.cn}
\equalcont{These authors contributed equally to this work.}

\author[1,3]{\fnm{Zhiwen} \sur{Xue}}\email{xuezhiwen@stu.xjtu.edu.cn}


\author*[1,3]{\fnm{Chunfeng} \sur{Lian}}\email{chunfeng.lian@xjtu.edu.cn}

\author*[2,3]{\fnm{Jianhua} \sur{Ma}}\email{jhma@xjtu.edu.cn}

\affil[1]{\orgdiv{School of Mathematics and Statistics}, \orgname{Xi'an Jiaotong University}, \orgaddress{\street{No.28, Xianning West Road}, \city{Xi'an}, \postcode{710049}, \state{Shaanxi}, \country{China}}}

\affil[2]{\orgdiv{Key Laboratory of Biomedical Information Engineering of Ministry of Education, School of Life Science and Technology}, \orgname{Xi'an Jiaotong University}, \orgaddress{\street{No.28, Xianning West Road}, \city{Xi'an}, \postcode{710049}, \state{Shaanxi}, \country{China}}}

\affil[3]{\orgdiv{Research Center for Intelligent Medical Equipment and Devices (IMED)}, \orgname{Xi'an Jiaotong University}, \orgaddress{\street{No.28, Xianning West Road}, \city{Xi'an}, \postcode{710049}, \state{Shaanxi}, \country{China}}}


\abstract{
Magnetic Resonance Imaging (MRI) is a cornerstone in medicine and healthcare but suffers from long acquisition times. Traditional accelerated MRI methods optimize for generic image quality, lacking adaptability for specific clinical tasks. To address this, we introduce PASS (Personalized, Anomaly-aware Sampling and reconStruction), an intelligent MRI framework that leverages a Vision-Language Model (VLM) to guide a deep unrolling network for task-oriented, fast imaging. PASS dynamically personalizes the imaging pipeline through three core contributions: (1) a deep unrolled reconstruction network derived from a physics-based MRI model; (2) a sampling module that generates patient-specific $k$-space trajectories; and (3) an anomaly-aware prior, extracted from a pretrained VLM, which steers both sampling and reconstruction toward clinically relevant regions. By integrating the high-level clinical reasoning of a VLM with an interpretable, physics-aware network, PASS achieves superior image quality across diverse anatomies, contrasts, anomalies, and acceleration factors. This enhancement directly translates to improvements in downstream diagnostic tasks, including fine-grained anomaly detection, localization, and diagnosis.
}

\maketitle

\section{Introduction}\label{sec1}


Magnetic resonance imaging (MRI) provides unparalleled soft-tissue contrast and non-invasive insight into morphology, function, and pathology, establishing it as a cornerstone of modern medical diagnosis and treatment planning~\cite{geraldes2018current}. However, its clinical utility is hampered by inherently prolonged acquisition times, which can induce patient discomfort, increase motion artifacts, and limit scanner throughput.

Consequently, significant research efforts aim to accelerate MRI by enhancing both physical hardware~\cite{griswold2002generalized} (e.g., multi-coil parallel imaging) and signal processing techniques for $k$-space undersampling and reconstruction~\cite{lustig2007sparse,hosny2018artificial,lingala2011accelerated}. Reconstruction methods from undersampled data fall into two broad categories, i.e., model-based and learning-based approaches. Traditional model-based methods integrate the MRI physical forward model with hand-crafted anatomical priors, offering interpretability but often faltering at high acceleration factors due to their limited representational power~\cite{doneva2010compressed,jin2016general}.

Deep learning breakthroughs have since bifurcated into data-driven and model-driven paradigms. Purely data-driven methods employ complex, black-box neural networks for end-to-end mapping, achieving impressive performance at the cost of interpretability and a heavy dependence on extensive training data~\cite{zhu2018image,wang2020deep,hammernik2018learning,sriram2020end,yiasemis2022recurrent}. In contrast, model-driven learning method-specifically deep unrolling-embed the physics of the acquisition within the network architecture by unrolling iterative optimization algorithms~\cite{Aggarwal2018,sun2016deep,Zhang2018}. This synthesis offers a compelling balance: it retains the interpretability and grounding of model-based approaches while harnessing the superior learning capacity of deep neural networks.

Despite these advancements, a critical limitation persists: the overwhelming focus on optimizing global voxel-level image quality (e.g., Peak Signal-to-Noise Ratio, PSNR)~\cite{lyu2023m4raw,Zbontar2018}. 
This paradigm lacks the adaptability for personalized, task-aware imaging, ultimately neglecting the enhanced visualization of patient-specific clinically relevant anomalies that is paramount for diagnostic accuracy.

Emerging approaches seek to bridge this gap by integrating downstream task information (e.g, segmentation) directly into the reconstruction process, often via multi-task learning frameworks~\cite{acar2022segmentation,fan2018segmentation,morshuis2024segmentation,ebner2020automated,jeong2024most}. While these methods show promise for targeted imaging, their reliance on empirical network architectures hinders interpretability and generalization across diverse pathologies. Parallel efforts to learn optimized $k$-space sampling patterns are also limited~\cite{bahadir2020deep,Peng2022,Wang2022}; these patterns are typically static and population-averaged, failing to adapt to the unique pathological characteristics of individual patients. Consequently, the ability to dynamically personalize the entire imaging pipeline (from acquisition to reconstruction) for specific clinical tasks remains an open challenge.

\begin{figure}[t]
	\centering
	\includegraphics[width=\textwidth]{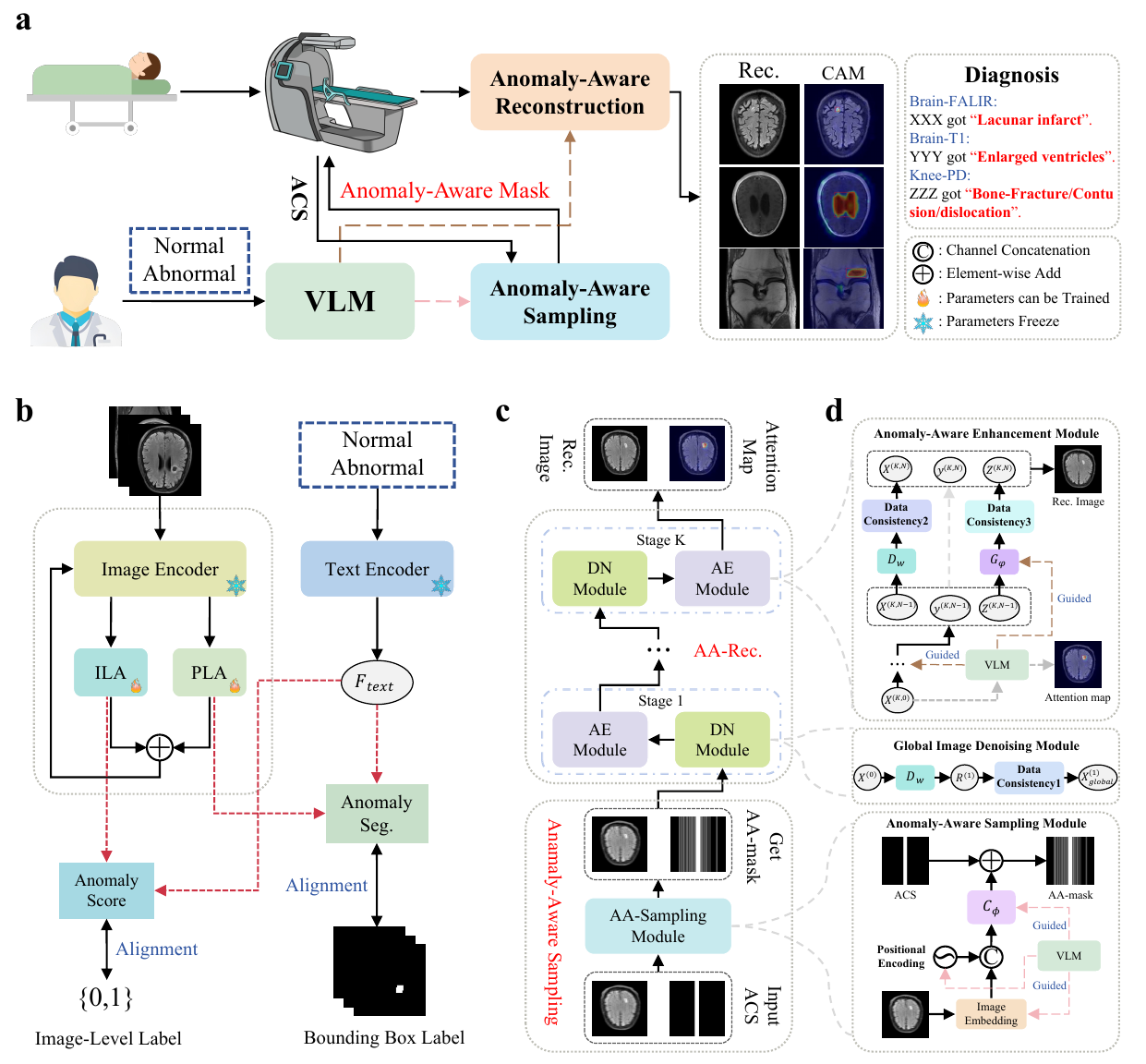}
	\caption{{PASS is a closed-loop system for personalized MRI. It introduces a paradigm where a fine-tuned VLM ($Net_{AD}$) guides the entire imaging pipeline from acquisition to reconstruction. As detailed in (a), the system overview, anomaly attention maps from the VLM direct a sampling network to acquire patient-specific $k$-space data, while simultaneously conditioning a deep unrolling reconstructor to enhance clinical features. The VLM's adaptation process (b) and the specialized architectures for sampling (c) and reconstruction (d) enable this co-design, ensuring the output is optimized for diagnostic relevance. Central $k$-space regions (autocalibration signals, ACS) provide a low-resolution prior, while image- and pixel-level adapters (ILA/PLA) enable precise VLM fine-tuning.}}
\label{fig1} 
\end{figure}

To address this open challenge, we propose PASS (Personalized, Anomaly-aware Sampling and reconStruction), a framework that fundamentally rethinks accelerated MRI by dynamically tailoring the entire imaging pipeline to patient-specific clinical needs. Our key insight is to leverage a pretrained Vision-Language Model (VLM) as a source of high-level, anomaly-aware prior knowledge to guide a physics-informed deep unrolling network. This synergy enables the joint optimization of adaptive, patient-specific $k$-space sampling trajectories and a conditioned reconstruction process, ensuring that both acquisition and reconstruction are explicitly oriented toward enhancing pathological features. Extensive validations across brain and knee MRI (spanning diverse contrasts, acceleration factors, and lesion types) demonstrate that PASS improves not only perceptual image quality but, more importantly, the accuracy of downstream diagnostic tasks such as fine-grained anomaly detection and diagnosis, all while maintaining the interpretability of a model-based foundation.

\section{Results}\label{sec2}
\subsection{Overview of the PASS Framework} 

PASS introduces a closed-loop pipeline for accelerated MRI that leverages a pretrained VLM as a source of high-level, anomaly-aware prior knowledge to dynamically tailor both acquisition and reconstruction to patient-specific clinical priorities (Fig.~\ref{fig1}a). The pipeline operates sequentially: an initial anomaly-aware sampling module first generates hardware-compatible, patient-specific $k$-space sampling trajectories focused on regions of potential diagnostic interest. This adaptive sampling is followed by a physics-informed, unrolled deep network---derived from a dedicated MRI reconstruction model with implicit regularizations---which iteratively refines the image through multiple cascaded stages. Each stage of this network integrates a global image denoising module to ensure overall data consistency and anatomical fidelity, and an anomaly-aware enhancement module that selectively sharpens pathological features under the guidance of the VLM (Fig.~\ref{fig1}c,d). The VLM itself, fine-tuned from CLIP~\cite{huang2024adapting,radford2021learning} in a multi-scale fashion using image-level labels and bounding boxes, provides a generalizable prior for diverse pathologies (Fig.~\ref{fig1}b). We rigorously evaluate PASS on the public fastMRI benchmark, augmented with anomaly labels from fastMRI+~\cite{Zhao2021}, under diverse knee and brain imaging protocols, assessing performance holistically through both quantitative image quality and downstream diagnostic utility.

\begin{figure}[t]
	\centering
\includegraphics[width=\textwidth]{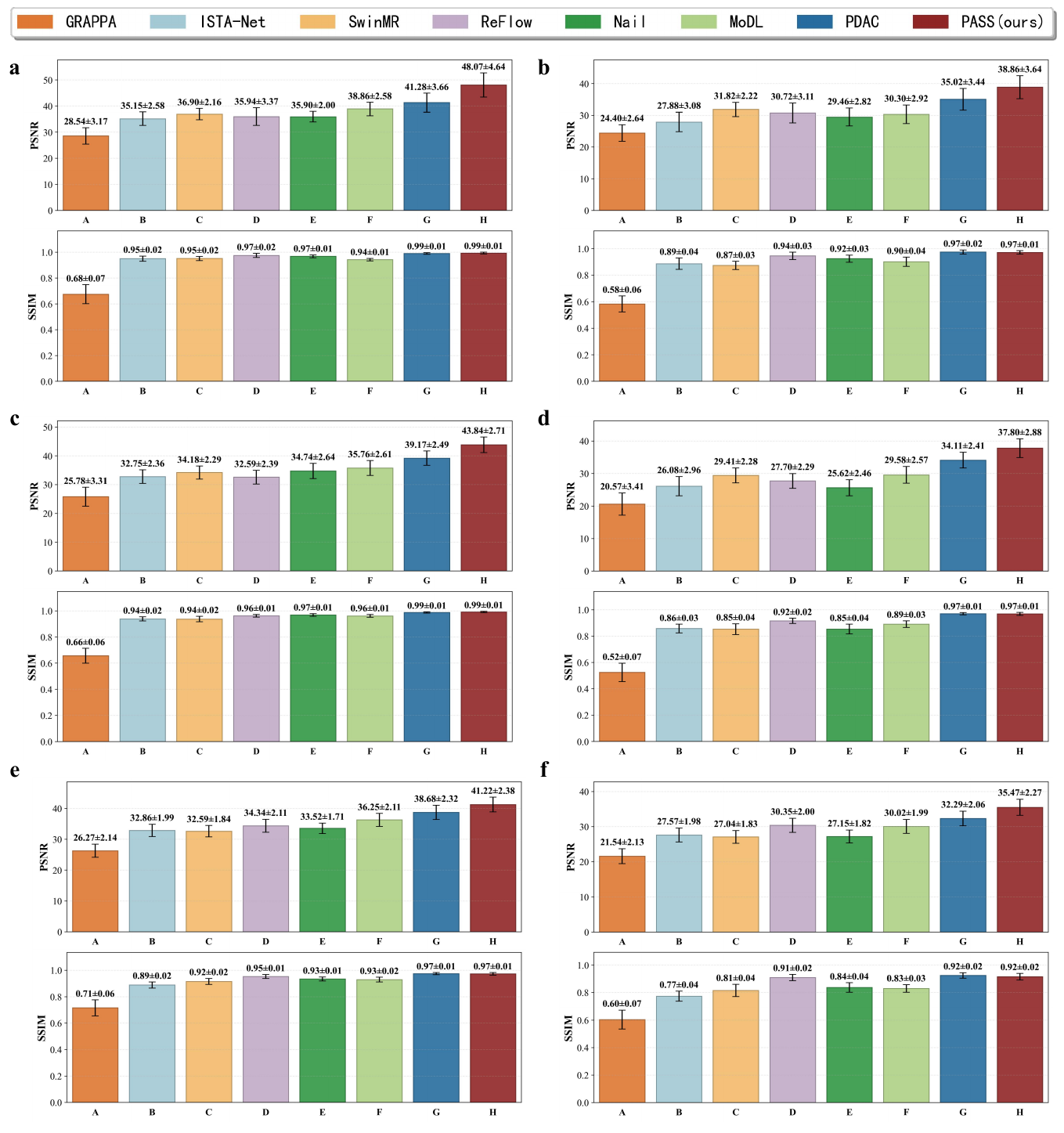}
	\caption{Quantitative comparison of reconstruction methods across datasets and acceleration factors. Left and right columns correspond to $4\times$ and $8\times$ acceleration, respectively. a–b, T1w brain, c–d, FLAIR brain, e–f, PD knee. PSNR and SSIM are reported for each method, with bars indicating mean values across the test set and error bars representing standard deviation.}
\label{fig2} 
\end{figure}

\subsection{Robust Performance Under Diverse Imaging Protocols}
PASS was rigorously evaluated on the fastMRI benchmark for brain (T1-weighted, T1w; Fluid-Attenuated Inversion Recovery, FLAIR) and knee (Proton Density-weighted, PD) imaging under low and high acceleration factors ($R=4\times,\, 8\times$). We compared its overall image quality, quantified by PSNR  and SSIM, against a comprehensive suite of representative methods: conventional model-based (i.e., GRAPPA~\cite{griswold2002generalized}), pure data-driven (i.e., SwinMR~\cite{huang2022swin}, Reflow~\cite{liu2022flow}, Nail~\cite{huang2024noise}, and PDAC~\cite{10657687}), and model-unrolled (i.e., ISTA-Net~\cite{Zhang2018} and MoDL~\cite{Aggarwal2018}) approaches. 

As summarized in Fig.~\ref{fig2}, PASS demonstrated consistent superiority, outperforming all competing methods by substantial margins across all anatomies, contrasts, and acceleration factors. Quantitative gains in mean PSNR were significant (e.g., $\Delta$ $+[6.79, 19.53]$ dB for brain T1w at $4\times$; and $\Delta$ $+[3.84,14.46]$ at $8\times$; similar robust improvements were observed for FLAIR and PD contrasts), underscoring its generalization capability.

\begin{figure}[]
	\centering
	\includegraphics[width=\textwidth]{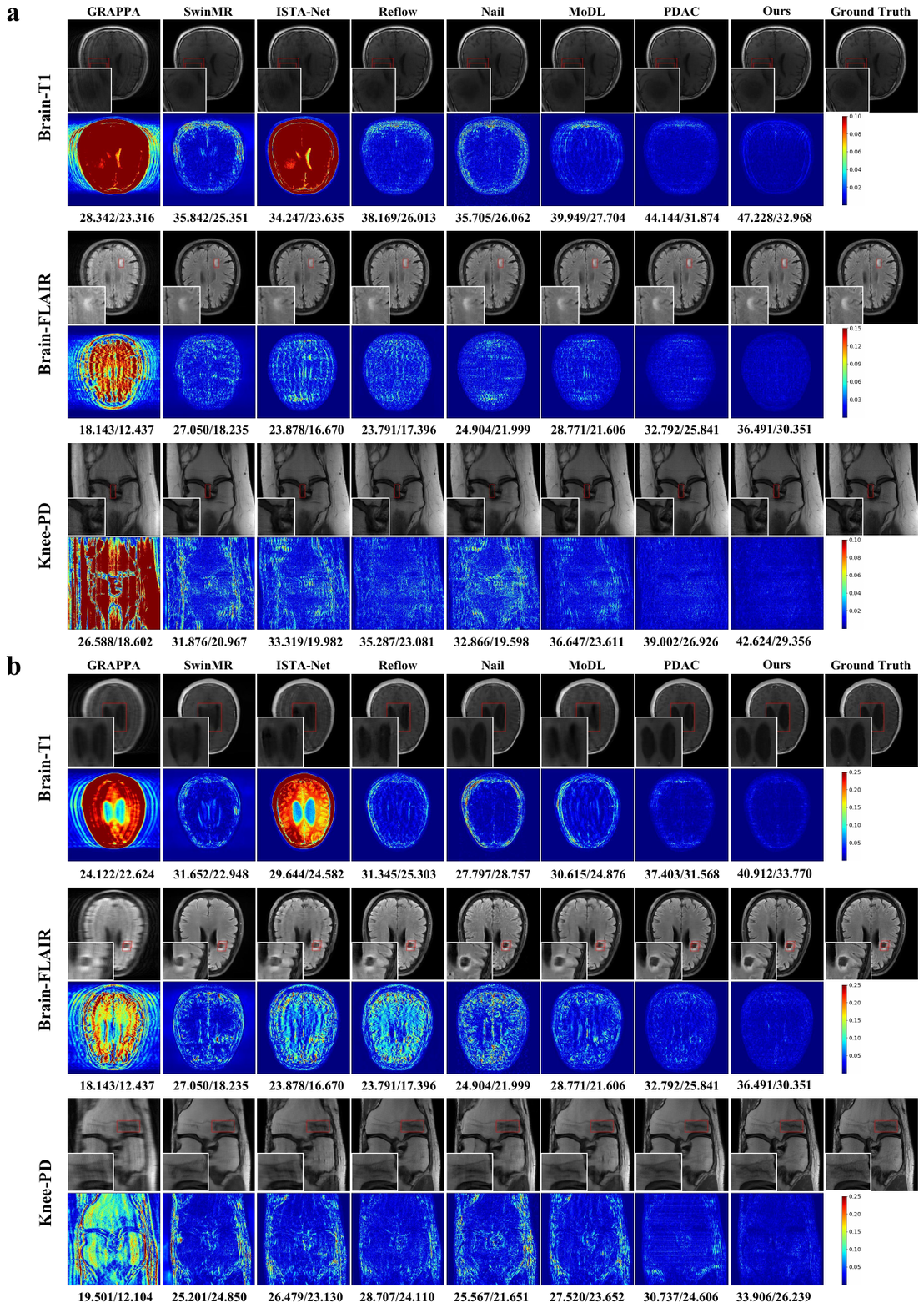}
	\caption{{Visual comparison of reconstructed images at $4\times$ (a) and $8\times$ (b) acceleration across different methods.
Representative results for T1w brain, FLAIR brain, and PD knee datasets. Each panel shows the reconstructed image and its corresponding error map relative to the fully sampled reference. PASS leads to sharper anatomical details and reduced residuals, particularly within pathological regions.}}
\label{fig3} 
\end{figure}

This quantitative superiority is corroborated by qualitative visual assessments (Fig.~\ref{fig3}). Reconstructions from PASS exhibit sharper anatomical details and reduced artifacts across diverse protocols. Crucially, zoomed-in views reveal that PASS provides enhanced visualization of pathological details. This consistent advantage, evident in both quantitative metrics and visual results, confirms the efficacy of its anomaly-aware design and its robust performance under highly variable clinical imaging scenarios.

\begin{figure}[]
	\centering
	\includegraphics[width=\textwidth]{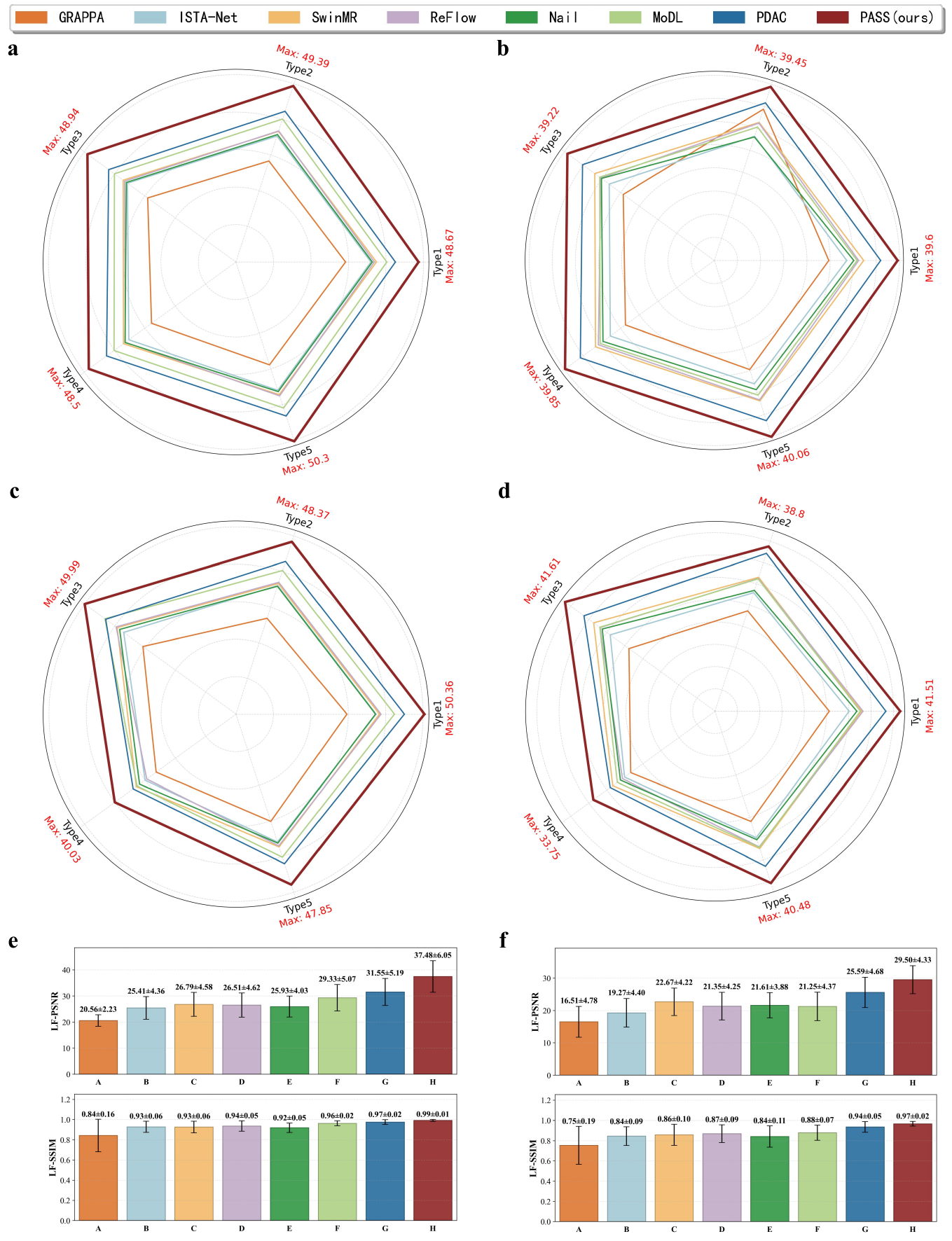}
	\caption{{Lesion-focused reconstruction evaluation.
Comparison of reconstruction quality across methods for T1w brain images. Left and right columns correspond to $4\times$ and $8\times$ acceleration, respectively. a-b, PSNR for the five most frequent lesions in the cohort: Type 1, Posttreatment change; Type 2, Mass; Type 3, Craniotomy; Type 4, Enlarged ventricles; Type 5, Edema. c-d, PSNR for five less frequent lesions in the cohort: Type 1, Resection cavity; Type 2, Normal variant; Type 3, Extra-axial mass; Type 4, Motion artifact; Type 5, Paranasal sinus opacification. e-f, Overall lesion-focused metrics (LF-PSNR and LF-SSIM) across all annotated regions. PASS consistently achieves superior fidelity in pathological areas.}}
\label{fig4} 
\end{figure}

\subsection{Enhanced Anomaly Imaging for Diverse Pathologies}

\begin{figure}[]
	\centering
	\includegraphics[width=\textwidth]{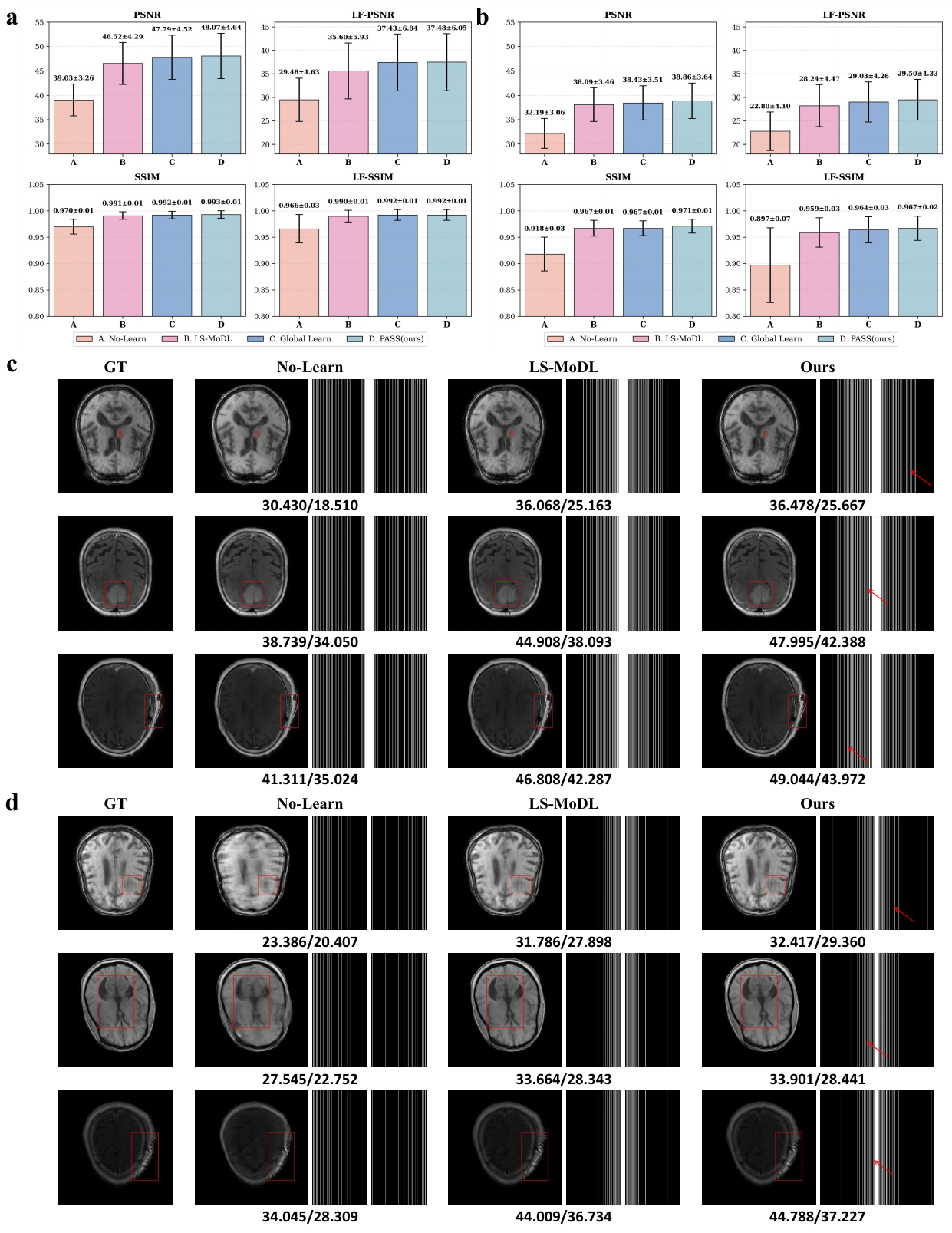}
	\caption{{Comparative analysis of sampling strategies for T1w brain images. Four strategies are evaluated: ``No-learn'', a fixed non-learnable mask, ``LS-MoDL'', a model-based baseline that jointly optimizes a reconstruction network and a learnable sampling mask, ``Global Learn'', a simplified variant of our framework employing LOUPE-based global sampling shared across all subjects without anomaly-aware sampling, and ``Ours'', the full PASS model integrating adaptive anomaly-aware sampling.
    a–b, Quantitative reconstruction results at $4\times$ and $8\times$, respectively, including PSNR, SSIM, LF-PSNR, and LF-SSIM.
c–d, Visual comparisons of reconstructed images and corresponding sampling masks at $4\times$ and $8\times$. Red arrows indicate regions where PASS adaptively increases sampling density in lesion-relevant $k$-space areas compared with other strategies.}}
\label{fig5} 
\end{figure}

To specifically evaluate the quality of pathological imaging, we quantified PASS's performance across both major and minor lesion types in the fastMRI dataset (Fig.~\ref{fig4}a-d, brain T1w; other contrasts/anatomies in Extended Fig.~1a-d and Fig.~2a-d). PASS significantly outperformed all competing methods in terms of PSNR across all these lesion categories, demonstrating stable and comparable performance agnostic to lesion frequency.

We further analyzed performance within localized lesion regions, defined by ground-truth bounding boxes in fastMRI+. PASS yielded substantial quantitative gains in these clinically critical areas (e.g., Fig.~\ref{fig4}e-f, PSNR improvements of $\Delta$ $+[5.93,16.92]$ dB and $+[3.91,12.99]$ dB for brain T1w at $4\times$ and $8\times$, respectively; other contrasts/anatomies in Extended Fig.~1e-f and Fig.~2e-f). These targeted improvements, consistent across diverse pathologies with varying frequencies, confirm the efficacy of the VLM-guided, anomaly-aware design and underscore its strong generalization capability to diverse and underrepresented pathological features.

\subsection{Adaptive, Anomaly-Aware $k$-Space Sampling}

To evaluate the efficacy of our adaptive sampling strategy, we conducted ablations by: (1) removing it (No-Learn, random sampling); (2) employing LOUPE-based global sampling shared across all subjects without anomaly-aware sampling (Global-Learn); and (3) comparing against LS-MoDL~\cite{Aggarwal2020}, a state-of-the-art method for joint sampling and reconstruction.

Results (Fig.~\ref{fig5}a-b, brain T1w; {Extended Fig.~3a-b, brain FLAIR; Extended Fig.~4a-b, brain PD}) demonstrate three key findings: first, learnable sampling is crucial, as all optimized strategies (Global-Learn, PASS) significantly outperformed random sampling (No-Learn), with PASS yielding $>7$ dB overal PSNR gains at both $4\times$ and $8\times$, confirming the broad value of acquisition optimization; second, task-oriented reconstruction matters, since both Global-Learn and PASS surpassed LS-MoDL, underscoring the importance of our dedicated architecture; and third, adaptivity provides a distinct advantage, as PASS consistently outperformed Global-Learn across all protocols, demonstrating the contribution of patient-specific sampling guided by the VLM's anomaly prior.
Qualitative analysis of the learned trajectories and corresponding reconstructions for diverse lesions (Fig.~\ref{fig5}c-d; {Extended Fig.~3c-d, Extended Fig.~4c-d}) provides further insight. PASS generated sampling patterns that are largely consistent (under the same anatomy/contrast) yet exhibit high-frequency specificities tailored to different pathologies. This adaptivity translates to the observed quantitative improvements in image quality, further validating the efficacy of our anomaly-aware sampling approach.

\begin{figure}[t]
	\centering
	\includegraphics[width=\textwidth]{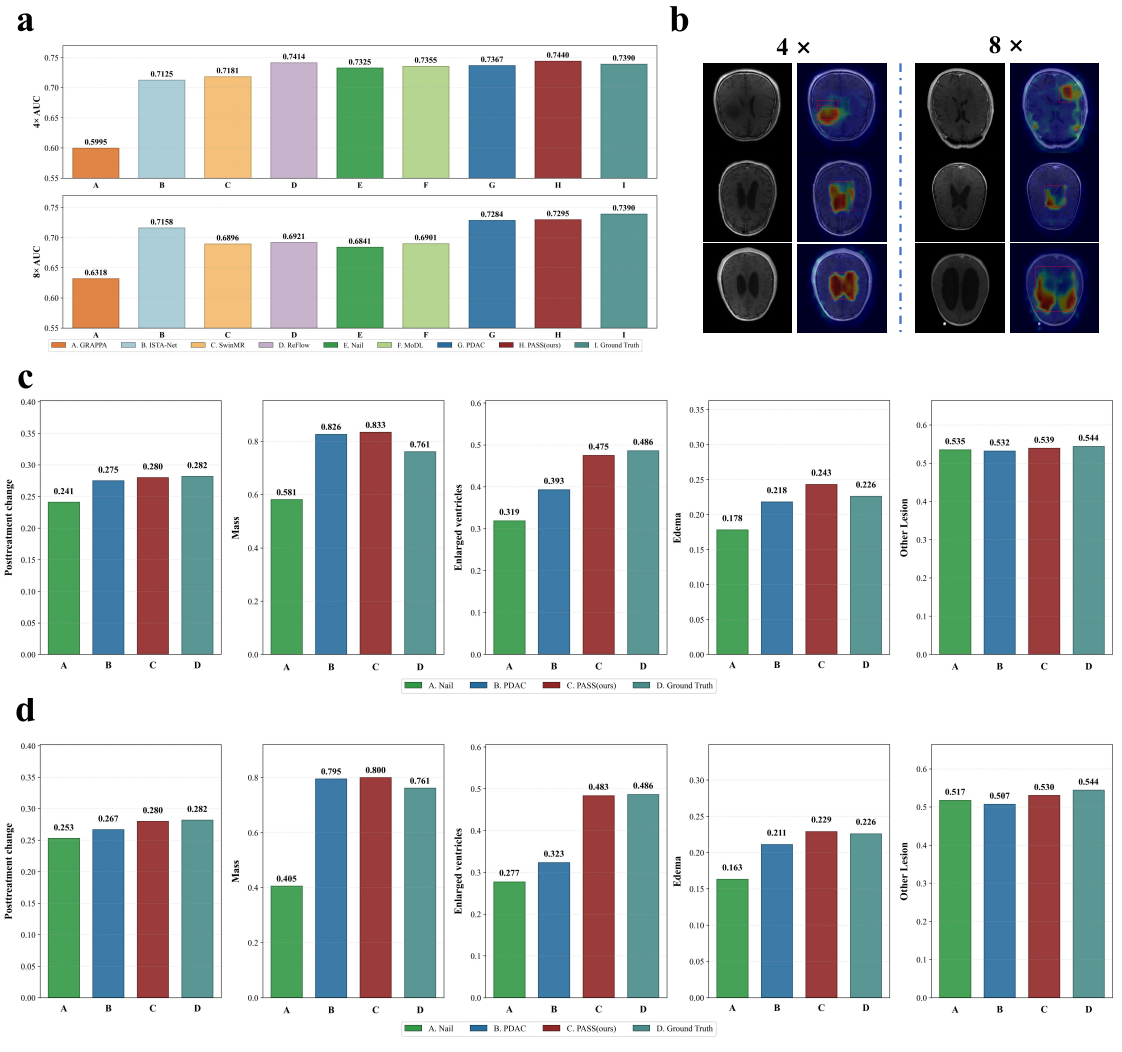}
	\caption{{Evaluation of downstream pathological tasks for T1w brain images. a, Anomaly detection AUC of a VLM-based detection model on reconstructed images generated by different methods under $4\times$ and $8\times$, evaluated across all lesions collectively.
b, Localization of pathological regions by the VLM-based detection model on PASS-reconstructed images.
c–d, Multi-label classification accuracy for individual lesion types using a fine-tuned CLIP image encoder with an MLP classifier, at $4\times$ (c) and $8\times$ (d), respectively.}}
\label{fig6} 
\end{figure}

\subsection{Improved Diagnostic Accuracy in Downstream Tasks}
To evaluate clinical utility, we quantified anomaly detection accuracy (AUC) on images reconstructed by different methods using a VLM-based detector~\cite{huang2024adapting} (Fig.~\ref{fig6}a; Extended Fig.~5a, Extended Fig.~6a). The results reveal two fundamental insights: first, superior voxel-level image quality (e.g., PSNR) does not guarantee improved diagnostic performance, exposing a critical gap between traditional MRI metrics and clinical application. For instance, in brain T1w, Reflow achieved lower PSNR than SwinMR ($35.94$ dB vs. $36.90$ dB at $4\times$) yet higher AUC ($74.14\%$ vs. $71.81\%$), underscoring this disconnect. Second, and most significantly, PASS achieved the highest overall AUC across different anatomies, contrasts, and acceleration factors, matching or even exceeding the performance on fully sampled ground-truth images. This demonstrates that a patient-tailored imaging pipeline directly enhances diagnostic precision. Qualitative inspection confirmed strong spatial agreement between PASS's fine-grained anomaly localization maps and ground-truth bounding boxes (Fig.~\ref{fig6}b; Extended Fig.~5b, Extended Fig.~6b).

To further dissect this pathology-centric advantage, we conducted a more challenging multi-class disease diagnosis task comparing the top-performing methods (Fig.~\ref{fig6}c-d, Extended Fig.~5c-d, Extended Fig.~6c-d). PASS consistently yielded the highest accuracy in differentiating pathological categories, maintaining this superiority at both $4\times$ and $8\times$ acceleration. Notably, for the ``Mass'' lesion type at $4\times$, PASS achieved an accuracy of $0.833$, significantly outperforming all competing methods and even surpassing the classification accuracy on the fully-sampled ground truth ($0.761$). This confirms that the explicit VLM-guided prior stabilizes fine-grained feature representations, producing reconstructions that are functionally superior for diagnostic tasks than those from state-of-the-art methods.

\section{Discussion}\label{sec3}
This study introduces PASS, a vision-language model (VLM)-guided deep unrolling framework that establishes a closed-loop, adaptive pipeline for fast MRI from acquisition to reconstruction. By leveraging a pretrained VLM to encode a generalizable prior over diverse pathologies, PASS dynamically tailors the imaging process: an anomaly-aware sampling module generates patient-specific $k$-space trajectories, which are then reconstructed by a physics-informed unrolled network optimized for pathological clarity. To our knowledge, this represents the first demonstration of a fully personalized framework for joint MRI acquisition and reconstruction, bridging cutting-edge foundation models with physics-aware deep learning to pioneer a new paradigm for MRI that simultaneously achieved interpretability and subject-specific adaptivity.

Conventional accelerated MRI methods are constrained by a fundamental limitation: their imaging protocols are typically designed for generic image quality, lacking the flexibility to adapt to patient-specific clinical presentations, particularly diverse pathologies. While recent approaches have incorporated downstream task regularizations to bridge this gap~\cite{ebner2020automated,jeong2024most,acar2022segmentation,fan2018segmentation,morshuis2024segmentation}, their generalizability remains hampered by static sampling learning and rigid multi-task architectures. In contrast, PASS introduces a fully adaptive pipeline, uniquely empowered by a VLM to dynamically guide both sampling and reconstruction. This innovation enables robust performance across a wide range of anatomies (brain, knee), contrasts (T1w, FLAIR, PD), acceleration factors ($4\times$, $8\times$), and---most critically---diverse lesion types (Figs.~\ref{fig2}-\ref{fig4}). Furthermore, extensive ablations studies under diverse experimental conditions confirm the efficacy of each component in PASS, notably the combination of VLM-guided, anomaly-aware sampling and reconstruction modules {(Extended Table~1).} Notably, PASS's advantages are most evident within pathological regions, even when global metrics appear comparable. For instance, on brain FLAIR at $8\times$ acceleration (Fig.~\ref{fig2}), PASS and the state-of-the-art PDAC achieved similar overall SSIM ($0.968$ vs. $0.969$), yet PASS significantly improved localized lesion quality (Extended Fig.~1f, bounding-box SSIM: $0.971$ vs. $0.942$). These results collectively affirm the superior flexibility and clinical generalizability of PASS for subject-specific, anomaly-aware MRI.

While $k$-space sampling patterns serve as a critical blueprint for accelerating MRI through strategic undersampling~\cite{bahadir2020deep,Peng2022,Wang2022}, conventional strategies, whether random or statically learned ~\cite{ravula2023optimizing,zhang2019reducing,lazarus2019sparkling,alush20203d,peng2022learning,wang2022b,weiss2019pilot}, inherently lack patient-specific adaptability. These rigid, one-size-fits-all approaches fail to prioritize clinically relevant regions or accommodate individual anatomical and pathological variations, ultimately limiting their diagnostic utility. To overcome this, PASS introduces an online, adaptive sampling framework that leverages the Auto-Calibrating Signal (ACS) to generate patient-specific trajectories in real-time, guided by the VLM's anomaly-aware prior. This enables targeted data acquisition that largely enhances image quality, particularly within pathological regions (Fig.~\ref{fig5}). Although this study primarily focuses on 1D sampling, preliminary investigations confirm the framework's extensibility to 2D sampling at high acceleration factors (e.g., $8\times$, $12\times$, $16\times$), where increased degrees of freedom yield even more diverse and subject-specific trajectories \textcolor{black}{(Extended Table~2)}. Together, this work establishes that adaptive, patient-specific sampling is not merely an incremental improvement but a fundamental paradigm shift for task-aware, accelerated MRI.

A pervasive limitation in current MRI paradigms is the methodological isolation of image reconstruction from downstream clinical utility. The field's prevailing focus on optimizing generic, voxel-level metrics (e.g., PSNR/SSIM) often overlooks a critical objective: whether the reconstructed images actually enhance performance in diagnostic tasks. Our results explicitly reveal this disconnect, demonstrating that superior global image quality dose not inherently translate to improved anomaly detection accuracy (Fig.~\ref{fig6}a). PASS directly addresses this gap by integrating diagnostic intent directly into the imaging pipeline. By leveraging the VLM to embed a generalizalbe, anomaly-aware prior, our framework ensure that both the acquisition and reconstruction processes are explicitly oriented toward enhancing pathological visibility and downstream diagnostic performance. Thus, PASS provides a feasible and effective blueprint for closing the loop between technical image quality and clinical value.

This study has several limitations that present avenues for future work. First, the VLM, fine-tuned from CLIP pretrained on nature images, was not fully optimized for the medical domain. A medicine-focused VLM pretrained on radiological data could enhance performance. Furthermore, our vision-language alignment currently prioritizes pathological information; incorporating imaging physics metadata (e.g., contrast weighting, echo times) could enable more targeted, protocol-specific enhancements. Second, the adaptive sampling module relies on the ACS as a reference. As the ACS contains primarily low-frequency data ($k$-space central lines), this may limit sensitivity to high-frequency pathological features. Future work could explore using the VLM to encode frequency-domain priors for ACS-based high-frequency outpainting to mitigate this. Third, evaluation was confined to the fastMRI benchmark. Expanding validation to include more diverse anatomies and contrast mechanisms is essential to further establish generalizability. Finally, all experiments were conducted retrospectively; prospective studies on clinical scanners are a critical next step for translating PASS into real-world practice.

In summary, by unifying vision-language intelligence with physics-informed deep learning, PASS provides a foundational step toward a new paradigm of clinically aware, patient-adaptive MRI.

\section{Methods}\label{sec4}

\subsection{PASS}
The PASS framework establishes a personalized paradigm for accelerated MRI that jointly optimizes acquisition and reconstruction with semantic guidance. It seamlessly integrates two core technical innovations: (i) a VLM-guided deep unrolling reconstructor that conditions image recovery on semantically-aware priors, and (ii) an adaptive, anomaly-aware sampling module that dynamically tailors k-space acquisition to individual patients. These two components form a closed diagnostic-acquisition loop that explicitly prioritizes regions of clinical relevance. The following sections detail the formulation and integration of these components within a unified optimization framework, as illustrated in Fig.~\ref{fig1}.


\subsubsection{Semantic Guidance via Vision-Language Models}
PASS pivots on a fine-tuned VLM, denoted as $Net_{AD}$, to generate a spatial attention map from intermediate image estimates, which pinpoints regions of potential clinical significance. By providing a dynamic, content-aware weighting, this map directly conditions the reconstruction network to concentrate on pathological features, effectively re-orienting the optimization goal from global fidelity to diagnostic relevance.

Our $Net_{AD}$ model was fine-tuned from a CLIP-based architecture for anomaly detection~\cite{huang2024adapting}. To retain the broad semantic knowledge of a pretrained CLIP backbone while adapting it to medical imaging, we freeze the CLIP encoders and insert lightweight adapters after the image encoder’s feature layers. Concretely, we introduce a Pixel-Level Adapter (PLA) to learn fine-grained localization and an Image-Level Adapter (ILA) to capture global pathology-relevant cues.
These adapters are trained with standard supervised signals (bounding-box supervision for PLA and image-level labels for ILA), producing attention maps that serve as the VLM-derived semantic priors used throughout PASS (Fig.~\ref{fig1}b). 

\subsubsection{VLM-Guided Deep Unrolling Reconstructor}
We cast personalized MRI reconstruction within an optimization-inspired, deep unrolling architecture. This design provides a principled and interpretable foundation for our learning-based approach. Specifically, we formulate the reconstruction as the minimization of an objective function that balances data fidelity, a general image prior, and a VLM-guided anomaly-aware regularizer:
\begin{equation}
X^* = \arg \min _{X, \Phi} \left\|A_{\Phi} X-Y\right\|_2^2+\mu_1\|X-D_w(X)\|_2^2\\+\mu_2H(Net_{AD}(X),X),
\label{eq1}
\end{equation}
where $Y$ are undersampled multi-coil k-space data and $A_{\Phi}=M_{\Phi}\mathcal{FS}$ denotes the forward model comprising coil sensitivities $\mathcal{S}$, the Fourier transform $\mathcal{F}$, and a learnable sampling mask $M_{\Phi}$. The term $D_w(\cdot)$ is a learned CNN denoiser that imposes a global image prior, and $H(\cdot, \cdot)$ encodes the VLM-guided, lesion-focused regularization that emphasizes reconstruction quality inside anomaly-attention regions.

We then address Eq.~\eqref{eq1} by unrolling its iterative optimization procedure into a multi-stage reconstruction network $R_{\Theta}$, parameterized by $\Theta$. The architecture of $R_{\Theta}$ explicitly mirrors the update rules of the underlying algorithm, offering both interpretability and flexibility, offering both interpretability and flexibility. Each stage $k$ of the network comprises two dedicated, learned modules that implement these updates:

\textbf{Denoising Module (DN):} DN first applies a CNN denoiser to the previous stage output $\mathbf{X}^{(k-1)}$ to suppress general noise and artifacts:
\begin{equation} 
R^{(k)}=D_w(X^{(k-1)}).
\end{equation}
This denoised intermediate is then refined via a data consistency update (labeled as \textit{Data Consistency 1} in Fig.~1), 
which enforces measurement fidelity by solving the following quadratic objective:
\begin{equation}
\mathbf{X}^{(k)}_{\text{global}} 
= \arg\min_{\mathbf{X}} 
\|A_{\Phi}\mathbf{X} - \mathbf{Y}\|_2^2 
+ \mu_1\|\mathbf{X} - R^{(k)}\|_2^2,
\end{equation}
which admits the closed-form solution 
\begin{equation}
X^{(k)}_{global}=(A_{\Phi}^{T}A_{\Phi}+\mu_1I)^{-1}(A_{\Phi}^{T}Y+\mu_1R^{(k)}),
\end{equation}
and is implemented efficiently via a conjugate-gradient (CG) solver. 
This procedure preserves physical fidelity while improving the overall global image quality.

\textbf{Personalized Anomaly-Aware Module (PA):} PA incorporates the VLM-derived prior to selectively refine regions of clinical interest. We implement this refinement by applying the Alternating Direction Method of Multipliers (ADMM) to solve the personalized regularization subproblem, which yields the following updates:
\begin{equation} 
\left\{
\begin{aligned}
    &X^{(k,n+1)}=\arg \min _{X} \left\|A_{\Phi} X-Y\right\|_2^2+\mu_1\|X-D_w(X)\|_2^2+\rho\left\|X-Z^{( n)}+\frac{y^{( n)}}{\rho}\right\|_2^2 \\ 
    &Z^{(n+1)}=\arg \min _{Z} \mu_2H(Net_{AD}(X^{(k)}_{global}),Z)+\rho\left\|X^{(k,n+1)}-Z+\frac{y^{( n)}}{\rho}\right\|_2^2 \\ 
    &y^{( n+1)} = y^{( n)}+\rho(X^{(k,n+1)}-Z^{( n+1)}).\\ 
\end{aligned}
\right.
\label{eq6}   
\end{equation}
The $\mathbf{X}$-update is solved via a CG algorithm, corresponding to the Data Consistency 2 block in Fig.~1. 
The $\mathbf{Z}$-update is implemented as a learned proximal step:
\begin{equation}
\left\{
\begin{aligned}
&U^{(n+1)} = G_{\varphi}(\mathrm{Net}_{AD}(X^{(k)}_{\mathrm{global}}), X^{(k,n+1)}),\\
&Z^{(n+1)} = (\rho+\mu_2)^{-1}(\mu_2U^{(n+1)}+\rho X^{(k,n+1)}+y^{(n)}),
\end{aligned}
\right.
\label{eq7}
\end{equation}
where the $Net_{AD}$ first generates an anomaly attention map from $\mathbf{X}^{(k)}{\text{global}}$, which then conditions the refinement network $G{\varphi}$ to produce the semantically enhanced features $\mathbf{U}^{(n+1)}$. 
The update of $\mathbf{Z}^{(n+1)}$ in Eq.~\eqref{eq7} then constitutes the Data Consistency 3 operation in Fig.~1, ensuring localized, anomaly-aware consistency in the final reconstruction by integrating semantic priors with physical constraints.

Finally, the reconstruction network $\mathcal{R}_{\Theta}$ is trained end-to-end with a composite loss:
\begin{equation}
\mathcal{L}_{\text{Rec}}=\gamma_1\|X^{(K)}_{global}-X^{(gt)}\|_2^2+\gamma_2\|X^{(K)}-X^{(gt)}\|_2^2+\gamma_3\|map*(X^{(K)}-X^{(gt)})\|_2^2.
\label{eq10}
\end{equation}
where $map=Net_{AD}(X_{global}^{(K)})$ denotes the VLM-derived attention map and $\mathbf{X}^{(gt)}$ the fully-sampled ground truth. 
The first two terms ensure accurate global reconstruction, while the final term emphasizes fidelity within pathology-relevant regions highlighted by the attention map.

\subsubsection{Adaptive and Anomaly-Aware Sampling}
The PASS framework closes the loop between acquisition and diagnosis by learning a patient-adaptive sampling mask $\mathbf{M}_{\Phi}$, optimized for specific clinical findings.  This is achieved via a two-stage optimization strategy designed to disentangle general image quality from diagnostic specificity.

\textbf{Stage 1: Establishing a Population-Level Prior.} The framework first learns a robust, general-purpose reconstructor $R_{\Theta}$ alongside a baseline probabilistic sampling mask $M_{\Phi}$ using the LOUPE paradigm~\cite{bahadir2020deep}. This stage establishes a foundational model and a population-optimal sampling strategy that guarantees general anatomical integrity.

\textbf{Stage 2: Personalized Anomaly-Aware Sampling.} With the reconstructor fixed, we introduce a lightweight anomaly-aware sampling network (AS) to personalize the acquisition. The AS network generates a high-frequency sampling component $M_{\Phi}^{2}$, which is combined with the fixed central k-space mask $M^{1}$ to form the complete mask 
$M_{\Phi}$ = $M^{1}$ + $M_{\Phi}^{2}$, where $M^{1}$ remains the fixed, fully-sampled central k-space, guaranteeing structural coherence. Conditioned on a low-resolution prior image, derived from the fully-sampled central k-space region defined by $\mathbf{M}^1$ (autocalibration signals), and sinusoidal coordinate embeddings, the AS network learns to allocate sampling points to k-space regions most informative for revealing patient-specific anomalies.

\textbf{A Diagnostic Feedback Loop.} The learning signal for this personalized sampler originates from a dedicated feedback loop. The undersampled k-space, acquired using the dynamic mask $M_{\Phi}$, is reconstructed by the fixed $R_{\Theta}$. The sampler then improves by minimizing a loss function that incorporates two competing objectives: global image fidelity and a novel anomaly-specific k-space consistency term. 
\begin{equation}
\mathcal{L}_{\text{mask}} = \|\mathbf{\textit{X}}^{(K)} - \mathbf{\textit{X}}^{(gt)}\|_2^2 + \lambda \left\|\mathbf{\textit{map}} \odot \left(\mathcal{F}^H(\mathbf{\textit{M}}_{\Phi}^{2} \odot \mathcal{F}(\mathbf{\textit{X}}^{(K)})) - \mathcal{F}^H(\mathbf{\textit{M}}_{\Phi}^{2} \odot \mathcal{F}(\mathbf{\textit{X}}^{(gt)}))\right)\right\|_2^2,
\label{eq:sampling_loss}
\end{equation}
This latter term directly penalizes discrepancies in the high-frequency k-space data within VLM-identified regions, directly teaching the sampler to preserve high-frequency information critical for diagnosing the patient's specific condition. Here $map$ denotes the attention map obtained from the anomaly detection network at the $K$ stage, i.e., $map=Net_{AD}(X_{global}^{(K)})$. This results in achieving a patient-specific balance between scan speed and diagnostic accuracy.

\subsubsection{Implementation Details} 
Our method is implemented on the PyTorch platform, optimized using Adam with a learning rate of 1e-3, a batch size of 8, and trained for 300 epochs.The network is unrolled for three iterations, with parameters shared across stages to balance model capacity and efficiency. The network model employs two separate input and output channels to represent the real and imaginary parts of the complex-valued MRI image. For sampling masks, we adopt both one-dimensional (1D) and two-dimensional (2D) undersampling strategies. In the 1D mask, the ratio between low- and high-frequency components follows the configuration of the fastMRI benchmark [24], adjusted for different acceleration factors. For the 2D mask, the low-frequency region is defined as a central $0.1 \times 0.1$ square in k-space, while high-frequency regions are randomly sampled according to the target acceleration factor. For the VLM-guided anomaly detection model~\cite{huang2024adapting}, we fine-tune it on each dataset using 12 annotated cases that contain both pixel-level bounding boxes and image-level lesion labels. After convergence, the model parameters are frozen and subsequently used to provide anomaly-aware priors during both sampling and reconstruction.
\subsection{Competing Methods}
We compared our proposed method against a range of representative MRI reconstruction approaches, covering traditional, model-based, transformer-based, and diffusion-based paradigms.

\begin{enumerate}[1)]
\item \textbf{GRAPPA}~\cite{griswold2002generalized}: A conventional parallel imaging technique that exploits inter-coil correlations to interpolate missing k-space data. Following the standard configuration, the interpolation kernel was set to $5\times5$, and the number of ACS lines was adjusted based on the acceleration factor (i.e., $8\%$ and $4\%$ of total lines for $4\times$ and $8\times$ acceleration, respectively).

\item \textbf{ISTA-Net}~\cite{Zhang2018}: A model-driven deep network unrolled from the Iterative Shrinkage–Thresholding Algorithm (ISTA), which learns the regularization and step-size parameters from data. We followed the authors’ settings with five unrolled stages, a learning rate of 0.001, and 100 training epochs.

\item \textbf{MoDL}~\cite{Aggarwal2018}: A model-based deep learning paradigm that alternates between CNN-based denoising and physics-consistent data consistency updates. The network was trained for three unrolled iterations with a learning rate of 0.001 over 100 epochs.

\item \textbf{SwinMR}~\cite{huang2022swin}: A transformer-based approach employing Swin Transformer blocks for data-driven MRI reconstruction. All hyperparameters followed the authors’ default configuration.

\item \textbf{Reflow}~\cite{liu2022flow}: A diffusion-based generative framework enabling one-step MRI reconstruction through fast deterministic sampling. The denoising network adopts a U-Net backbone, consistent with the official implementation.

\item \textbf{Nail}~\cite{huang2024noise}: A diffusion model specifically designed for MRI reconstruction that enforces data consistency at every denoising step to ensure physics-constrained generation. The original hyperparameters were retained.

\item \textbf{PDAC}~\cite{10657687}: A degradation-decomposition strategy that progressively restores severely aliased MR images via multiple moderate reconstruction steps. We adopted the implementation based on HUMUS-Net using the authors’ recommended settings.

\item \textbf{LS-MoDL}~\cite{Aggarwal2020}: To investigate the impact of learnable k-space sampling, we extended MoDL with the LOUPE optimization strategy. In this variant, the sampling mask is jointly optimized with the reconstruction network in an end-to-end manner and shared across all slices. The design respects the physical constraints of MRI acquisition, providing a fair comparison between static and adaptive sampling strategies.
\end{enumerate}

\subsection{Fine-Grained Anomaly Detection and Diagnosis}
To quantitatively evaluate the capability of reconstructed images in preserving clinically relevant lesions, we designed specific downstream diagnostic tasks. First, we evaluated the effectiveness of the VLM-derived attention maps by computing the area under the curve (AUC) of a fine-tuned VLM model on reconstructed images generated by different methods. This evaluation measures how well the attention-guided prior highlights regions of potential pathology. Second, we constructed a multi-label lesion classification task to assess reconstruction fidelity across lesion types. The classifier consists of a frozen CLIP image encoder for feature extraction and a MLP for classification. For each dataset, four common lesion types in the cohort were selected, with an additional category representing other lesions. The classifier is trained on fully-sampled ground truth images and then applied to reconstructed images from each method, enabling quantitative evaluation using standard metrics, i.e., accuracy (ACC).  These tasks provide lesion-specific benchmarks to systematically compare reconstruction methods in terms of their ability to preserve pathological details.


\subsection{Data Preprocessing and Evaluation}

Experiments were conducted on multi-coil brain and knee datasets from the publicly available fastMRI repository~\cite{Zbontar2018}. Coil sensitivity maps were estimated using the ESPIRiT algorithm~\cite{uecker2014espirit}. Brain scans included T1-weighted and FLAIR contrasts, while knee scans consisted of proton-density (PD) images.
To account for variable raw k-space dimensions in the brain dataset, k-space were first transformed to the image domain and uniformly cropped to 320 × 320 pixels. Corresponding k-space data and coil sensitivity maps were were then simulated from these cropped images. In contrast, knee scans were acquired at a standardized $320 \times 320$ resolution, consistent with the original acquisition protocol, and required no further preprocessing.

Pathological annotations were sourced from the fastMRI+ dataset, which provides both lesion types and bounding box coordinates. For brain scans, image-level labels were derived from study-level annotations~\cite{Zhao2021}. Slices belonging to positively labeled studies were assigned a binary label indicating the presence of lesions, although explicit bounding boxes were not available for these slices.

Model performance was evaluated using conventional metrics, including peak signal-to-noise ratio (PSNR) and structural similarity index (SSIM). To specifically assess quality in clinically relevant regions, we further introduced lesion-focused metrics, LF-PSNR and LF-SSIM, computed exclusively within annotated lesion areas.



\section{Data Availability}

The multi-coil knee and brain MRI data used in this study were sourced from the public fastMRI dataset (https://fastmri.org/). The corresponding pathological annotations were obtained from the fastMRI+ repository (https://github.com/microsoft/fastmri-plus). All custom-processed data, including preprocessed images and coil sensitivity maps, are publicly available on Zenodo (https://zenodo.org/records/PASS) and GitHub (https://github.com/ladderlab-xjtu/PASS).

\section{Code Availability}

The implementation of PASS, including the model architecture, training procedures, and evaluation framework, is publicly available at https://github.com/ladderlab-xjtu/PASS. A permanent archive of the codebase will be deposited on Zenodo (https://zenodo.org/records/PASS) to ensure long-term reproducibility. The framework is implemented in PyTorch using exclusively open-source dependencies.



\section{Acknowledgments}
The authors acknowledge the funding of the the National Natural Science Foundation of China (Nos.~T2522028 and 12326616), National Key R\&D Program of China (No.~2022YFA1004200), Natural Science Basic Research Program of Shaanxi (No.~2024JC-TBZC-09), and Shaanxi Provincial Key Industrial Innovation Chain Project (No.~2024SF-ZDCYL-02-10).


\section{Competing Interests}
The authors have no conflict of interest to declare.

\end{document}


\section{Extended Data}

To investigate the impact of key components of our method, we conducted ablation studies by separately handling them to examine their influence on reconstruction performance. As show in Table~\ref{tab1}.
1) PASS/$\mathrm{\Phi}$,PA: Removed the personalized anomaly-aware module
 and all learnable styles, corresponding to the setup mentioned in MoDL. 
2) PASS/$\mathrm{\Phi}$: Removed all learnable styles. 
3) PASS/PA,AS: Removed both the personalized anomaly-aware module and the anomaly-aware sampling module. Set as LOUPE samping strategy to learn global samping mask. Corresponding to the setup mentioned in LS-MoDL. 
4) PASS/AS: Removed the anomaly-aware sampling module while keeping the sampling strategy as LOUPE.

\begin{table*}[htbp]
  \centering
  \caption{Ablation studies to check the effect of each key component of our PASS.}
  \label{tab1}
  \resizebox{\textwidth}{!}{
  \begin{tabular}{cccccccccc}
		\toprule
		\multicolumn{2}{c}{\textbf{1D}} & \multicolumn{4}{c}{$4\times$}    & \multicolumn{4}{c}{$8\times$} \\
		\cmidrule(lr){1-2} \cmidrule(lr){3-6} \cmidrule(lr){7-10}
        Contrast & Method & PSNR & LF-PSNR & SSIM & LF-SSIM & PSNR & LF-PSNR & SSIM & LF-SSIM \\
  
		\midrule
        \multirow{5}{*}{Brain-FLAIR}
  &PASS/$\mathrm{\phi}$,PA       & 35.76$\pm$2.61   &26.93$\pm$3.98          & 0.961$\pm$0.011  &0.964$\pm$0.031              & 29.58$\pm$2.57  &21.48$\pm$4.33          & 0.890$\pm$0.025   &0.896$\pm$0.079          \\
  
  &PASS/$\mathrm{\phi}$             & 35.95$\pm$2.68  &27.15$\pm$3.99          & 0.960$\pm$0.014  &0.962$\pm$0.035                  & 30.61$\pm$2.61  &22.00$\pm$3.93          & 0.898$\pm$0.038  &0.910$\pm$0.071          \\
  
  &PASS/PA,AS          & 43.38$\pm$2.87   &34.45$\pm$4.08          & 0.990$\pm$0.006  &0.992$\pm$0.009               & 35.80$\pm$2.56   &27.07$\pm$4.23          & 0.959$\pm$0.012   &0.961$\pm$0.033          \\
  
  &PASS/AS     & 43.57$\pm$2.68   &34.56$\pm$4.26          & 0.991$\pm$0.005   &0.992$\pm$0.009         & 36.92$\pm$2.96  &28.15$\pm$4.28          & 0.966$\pm$0.012   &0.969$\pm$0.028          \\
  
  
  &\textbf{PASS} & \textbf{43.84$\pm$2.71} &\textbf{34.97$\pm$3.95} & \textbf{0.991$\pm$0.005} &\textbf{0.992$\pm$0.009} & \textbf{37.80$\pm$2.88} &\textbf{28.54$\pm$4.21} & \textbf{0.968$\pm$0.011} &\textbf{0.971$\pm$0.027} \\  
  \midrule[0.5pt]

\multirow{5}{*}{Brain-T1w} 
  &PASS/$\mathrm{\phi}$,PA       & 38.86$\pm$2.58 & 29.33$\pm$5.07 & 0.942$\pm$0.010 & 0.963$\pm$0.024 &
30.30$\pm$2.92 & 21.25$\pm$4.37 & 0.900$\pm$0.035 & 0.879$\pm$0.075          \\
  
  &PASS/$\mathrm{\phi}$             & 39.03$\pm$3.26 & 29.48$\pm$4.63 & 0.970$\pm$0.014 & 0.966$\pm$0.027 &
32.19$\pm$3.06 & 22.80$\pm$4.10 & 0.918$\pm$0.032 & 0.897$\pm$0.071          \\
  
  &PASS/PA,AS          & 46.52$\pm$4.29 & 35.60$\pm$5.93 & 0.991$\pm$0.007 & 0.990$\pm$0.011 &
38.09$\pm$3.46 & 28.24$\pm$4.47 & 0.967$\pm$0.015 & 0.959$\pm$0.028 \\
  
  &PASS/AS     & 47.79$\pm$4.52 & 37.43$\pm$6.04 & 0.992$\pm$0.007 & 0.992$\pm$0.010 &
38.43$\pm$3.51 & 29.03$\pm$4.26 & 0.967$\pm$0.014 & 0.964$\pm$0.025          \\
  
  
&\textbf{PASS} &
\textbf{48.07$\pm$4.64} &
\textbf{37.48$\pm$6.05} &
\textbf{0.993$\pm$0.007} &
\textbf{0.992$\pm$0.010} &
\textbf{38.86$\pm$3.64} &
\textbf{29.50$\pm$4.33} &
\textbf{0.971$\pm$0.013} &
\textbf{0.967$\pm$0.023} \\  
  \midrule[0.5pt]

\multirow{5}{*}{Knee-PD} 
  &PASS/$\mathrm{\phi}$,PA       & 36.25$\pm$2.11 &26.50$\pm$2.19          & 0.929$\pm$0.018  &0.951$\pm$0.023              & 30.02$\pm$1.99   &22.93$\pm$2.68          & 0.829$\pm$0.028   &0.909$\pm$0.050          \\
  
  &PASS/$\mathrm{\phi}$            & 36.81$\pm$2.48   &27.07$\pm$2.27          & 0.939$\pm$0.019   &0.956$\pm$0.021                  & 31.61$\pm$2.47   &23.27$\pm$2.85          & 0.865$\pm$0.034   &0.915$\pm$0.045          \\
  
  &PASS/PA,AS          & 40.27$\pm$2.31  &29.67$\pm$2.12          & 0.964$\pm$0.011  &0.973$\pm$0.013
  & 34.97$\pm$2.11   &25.56$\pm$2.14          & 0.907$\pm$0.024  &0.939$\pm$0.031          \\
  
  &PASS/AS    & 41.14$\pm$2.36    &30.44$\pm$2.13         & 0.972$\pm$0.010  &0.978$\pm$0.011         & 35.39$\pm$2.33   &26.17$\pm$2.08          & 0.916$\pm$0.023  &0.943$\pm$0.029          \\
  
  
  &\textbf{PASS} & \textbf{41.22$\pm$2.38} &\textbf{30.48$\pm$2.13} & \textbf{0.972$\pm$0.010} &\textbf{0.978$\pm$0.011} & \textbf{35.47$\pm$2.27} &\textbf{26.29$\pm$2.02} & \textbf{0.915$\pm$0.024} &\textbf{0.943$\pm$0.029} \\  
  \midrule[0.5pt]
  
  \end{tabular}
  }
\end{table*}

To further investigate the key components of our proposed sampling methodology in PASS,
we conducted ablation studies by separately handling them to examine their influence on reconstruction performance with the context of 2D samping.
The component ablation setup matches the description above, with results shown in Table~\ref{tab2}.

\begin{table*}[htbp]
	\centering
	\caption{Ablation studies on our PASS components with the context of 2D samping.}
	\label{tab2}
    \resizebox{\textwidth}{!}{
	\begin{tabular}{cccccccccc}
		\toprule
		\multicolumn{2}{c}{\textbf{2D}} & \multicolumn{4}{c}{Brain-FLAIR}    & \multicolumn{4}{c}{Knee-PD} \\
		\cmidrule(lr){1-2} \cmidrule(lr){3-6} \cmidrule(lr){7-10}
        Acceleration & Method & PSNR & LF-PSNR & SSIM & LF-SSIM & PSNR & LF-PSNR & SSIM & LF-SSIM \\
  
		\midrule
		\multirow{3}{*}{$8\times$} 
    & PASS/PA,AS & 44.94$\pm$3.00         & 35.80$\pm$4.47          & 0.992$\pm$0.004          & 0.994$\pm$0.007          
                 & 42.82$\pm$2.48          & 31.02$\pm$2.29          & 0.978$\pm$0.008          & 0.980$\pm$0.011 \\
	
    & PASS/AS  & 45.34$\pm$2.91 & 36.45$\pm$4.19 & 0.993$\pm$0.004 & 0.995$\pm$0.006  
               & 44.08$\pm$2.41 & 32.23$\pm$2.37 & 0.983$\pm$0.006 & 0.984$\pm$0.009 \\
		
& \textbf{PASS} 
& \textbf{45.77$\pm$3.05} 
& \textbf{36.80$\pm$4.19} 
& \textbf{0.994$\pm$0.004} 
& \textbf{0.995$\pm$0.006} 
& \textbf{44.19$\pm$2.38} 
& \textbf{32.32$\pm$2.36} 
& \textbf{0.983$\pm$0.006} 
& \textbf{0.984$\pm$0.009} \\
		
        \midrule[0.5pt]
		\multirow{3}{*}{$12\times$} 
    & PASS/PA,AS & 41.13$\pm$2.84         & 32.20$\pm$4.02          & 0.983$\pm$0.006          & 0.985$\pm$0.014          
                 &39.87$\pm$2.35          & 28.73$\pm$2.06          & 0.958$\pm$0.014          & 0.965$\pm$0.017 \\
 	
    & PASS/AS & 42.72$\pm$3.05 & 33.20$\pm$4.46 & 0.989$\pm$0.005 &  0.990$\pm$0.010  
              & 40.54$\pm$2.42 & 29.26$\pm$2.10 & 0.965$\pm$0.012 & 0.969$\pm$0.015  \\
		
& \textbf{PASS} 
& \textbf{43.18$\pm$3.08} 
& \textbf{34.16$\pm$4.55} 
& \textbf{0.990$\pm$0.005} 
& \textbf{0.991$\pm$0.010} 
& \textbf{40.68$\pm$2.47} 
& \textbf{29.36$\pm$2.12} 
& \textbf{0.966$\pm$0.012} 
& \textbf{0.970$\pm$0.015} \\
		
		\midrule[0.5pt]
    \multirow{3}{*}{$16\times$} 
    & PASS/PA,AS & 38.90$\pm$2.87         & 29.89$\pm$4.39          & 0.978$\pm$0.031          & 0.980$\pm$0.018          
                 &38.25$\pm$2.39          & 27.48$\pm$2.07          & 0.940$\pm$0.019          & 0.954$\pm$0.022 \\
    
    & PASS/AS & 39.28$\pm$2.86 & 30.50$\pm$3.86 & 0.977$\pm$0.008 &   0.981$\pm$0.018  
              & 39.04$\pm$2.51 & 28.03$\pm$2.11 & 0.952$\pm$0.016 & 0.960$\pm$0.019  \\
& \textbf{PASS}
& \textbf{39.66$\pm$2.72}
& \textbf{31.04$\pm$3.58}
& \textbf{0.978$\pm$0.008}
& \textbf{0.981$\pm$0.018}
& \textbf{39.13$\pm$2.54}
& \textbf{28.09$\pm$2.12}
& \textbf{0.953$\pm$0.017}
& \textbf{0.960$\pm$0.019} \\
	   
    \bottomrule
	\end{tabular}
}
\end{table*}

\begin{figure}[]
	\centering
	\includegraphics[width=\textwidth]{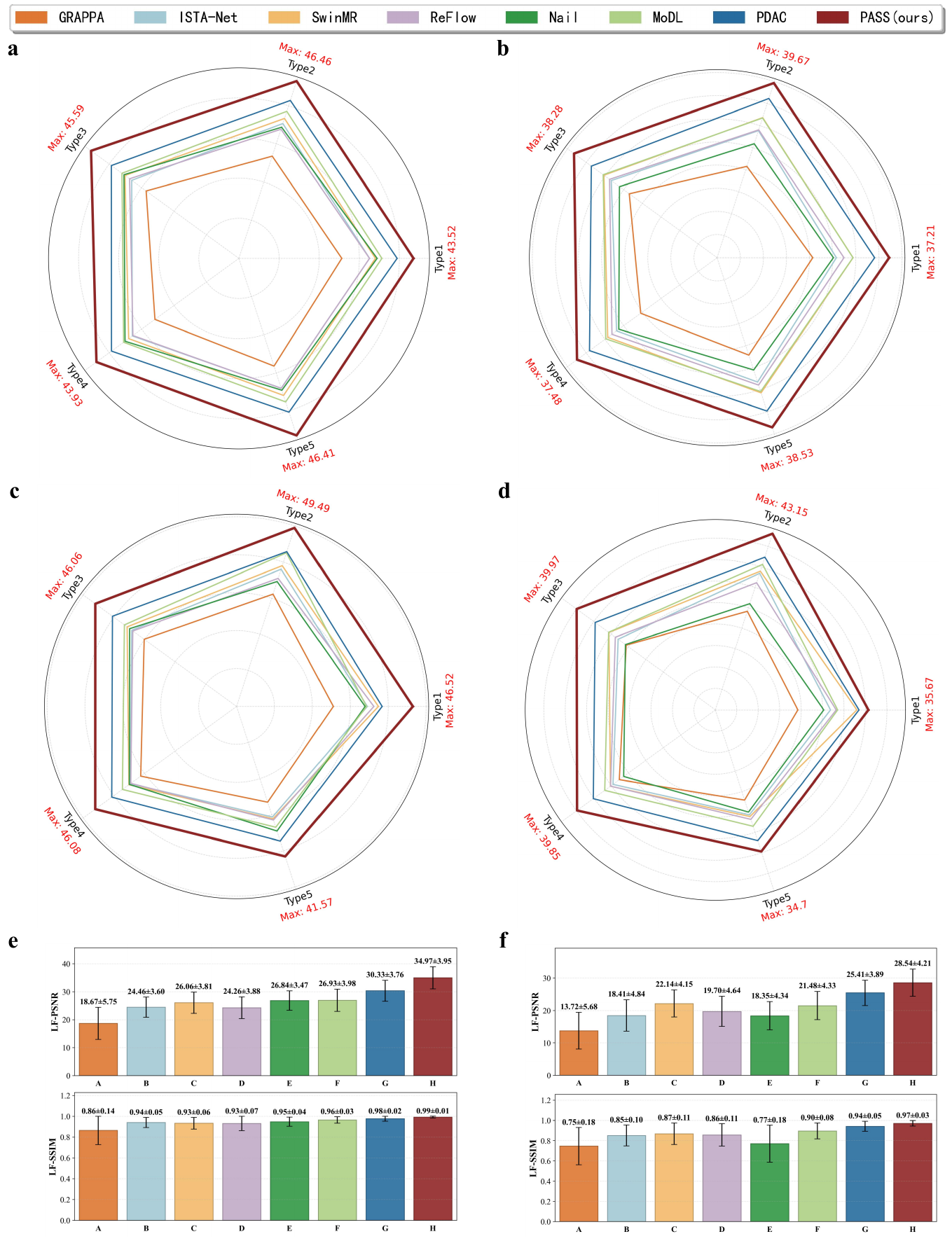}
	\caption{{Lesion-focused reconstruction evaluation.
Comparison of reconstruction quality across methods for FLAIR brain images. Left and right columns correspond to $4\times$ and $8\times$ acceleration, respectively. a-b, PSNR for the five most frequent lesions in the cohort: Type 1, nonspecific white matter lesion; Type 2, craniotomy; Type 3, encephalomalacia; Type 4, Global label: small vessel chronic white matter ischemic change; Type 5, posttreatment change. c-d, PSNR for five less frequent lesions: Type 1, dural thickening; Type 2, craniectomy with cranioplasty; Type 3, mass; Type 4, edema; Type 5, normal variant. e-f, Overall lesion-focused metrics (LF-PSNR and LF-SSIM) across all annotated regions. PASS consistently achieves superior fidelity in pathological areas.}}
\label{extend-fig1} 
\end{figure}

\begin{figure}[]
	\centering
	\includegraphics[width=\textwidth]{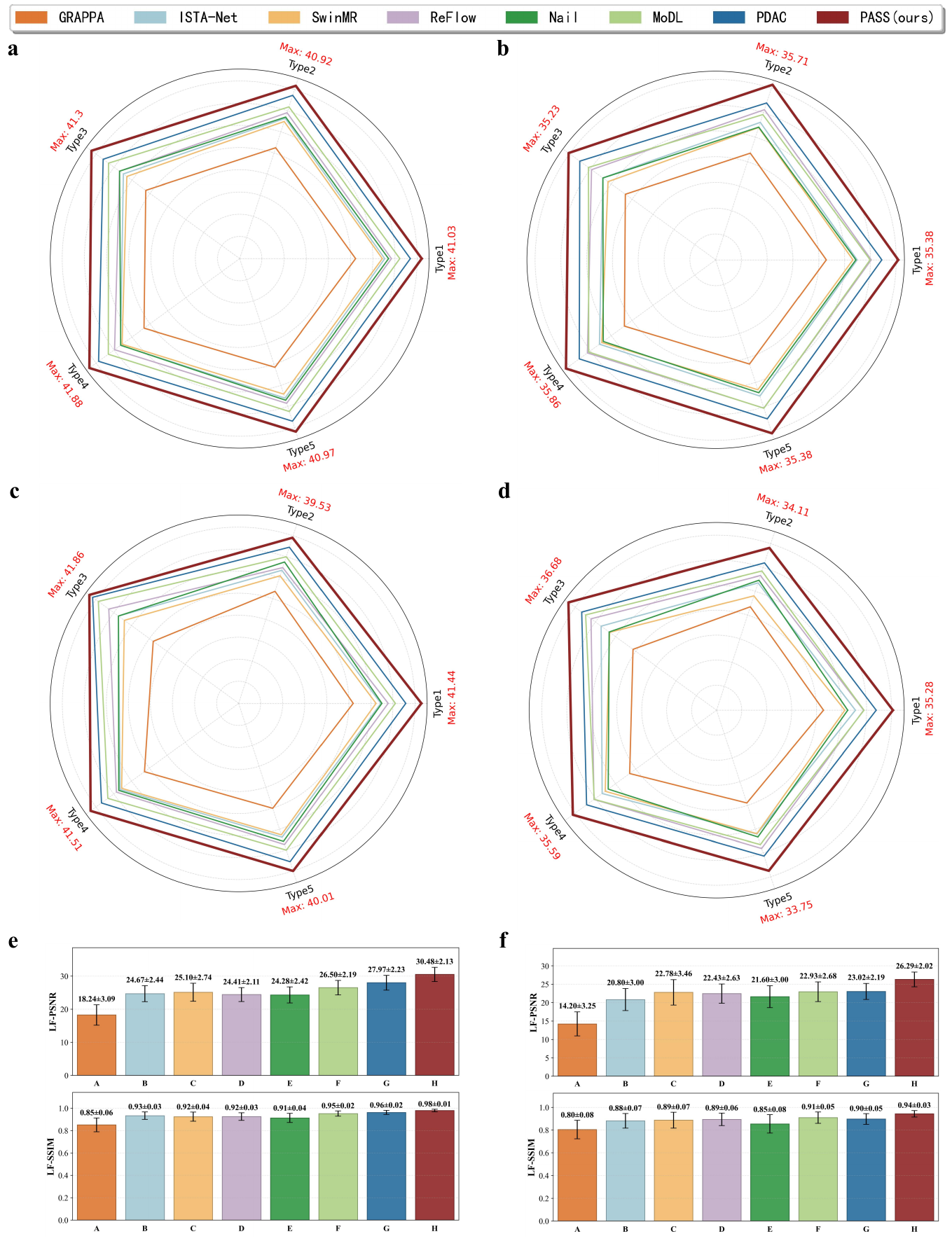}
	\caption{{Lesion-focused reconstruction evaluation.
Comparison of reconstruction quality across methods for PD Knee images. Left and right columns correspond to $4\times$ and $8\times$ acceleration, respectively. a-b, PSNR for the five most frequent lesions in the cohort: Type 1, Meniscus Tear; Type 2, Cartilage - Partial Thickness loss/defect; Type 3, Ligament - ACL Low Grade sprain; Type 4, Ligament - ACL High Grade Sprain; Type 5, Cartilage - Full Thickness loss/defect. c-d, PSNR for five less frequent lesions: Type 1, LCL Complex - Low-Mod Grade Sprain; Type 2, Soft Tissue Lesion; Type 3, Ligament - PCL Low-Mod grade sprain; Type 4, Bone- Subchondral edema; Type 5, Bone-Fracture/Contusion/dislocation. e-f, Overall lesion-focused metrics (LF-PSNR and LF-SSIM) across all annotated regions. PASS consistently achieves superior fidelity in pathological areas.}}
\label{extend-fig2} 
\end{figure}

\begin{figure}[]
	\centering
	\includegraphics[width=\textwidth]{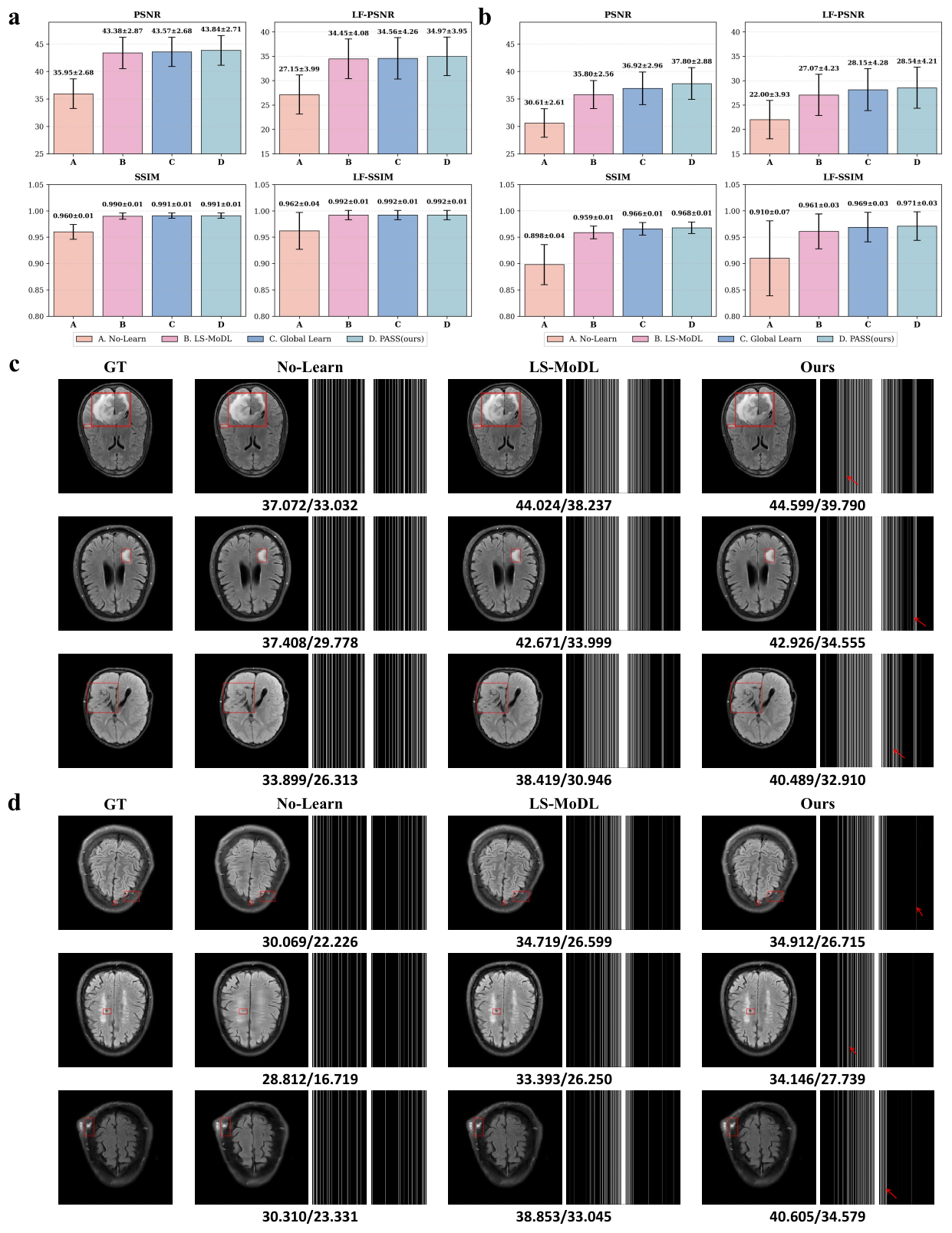}
	\caption{{Comparative analysis of sampling strategies for FLAIR brain images. Four strategies are evaluated, "No-learn": a fixed non-learnable mask, "LS-MoDL", a model-based baseline that jointly optimizes a reconstruction network and a learnable sampling mask, "Global Learn", a simplified variant of our framework employing LOUPE-based global sampling shared across all subjects without anomaly-aware sampling, and "Ours", the full PASS model integrating adaptive anomaly-aware sampling.
    a–b, Quantitative reconstruction results at $4\times$ and $8\times$ acceleration, respectively, including PSNR, SSIM, LF-PSNR, and LF-SSIM.
c–d, Visual comparisons of reconstructed images and corresponding sampling masks at $4\times$ and $8\times$ acceleration. Red arrows indicate regions where PASS adaptively increases sampling density in lesion-relevant k-space areas compared with other strategies.}}
\label{extend-fig3} 
\end{figure}

\begin{figure}[]
	\centering
	\includegraphics[width=\textwidth]{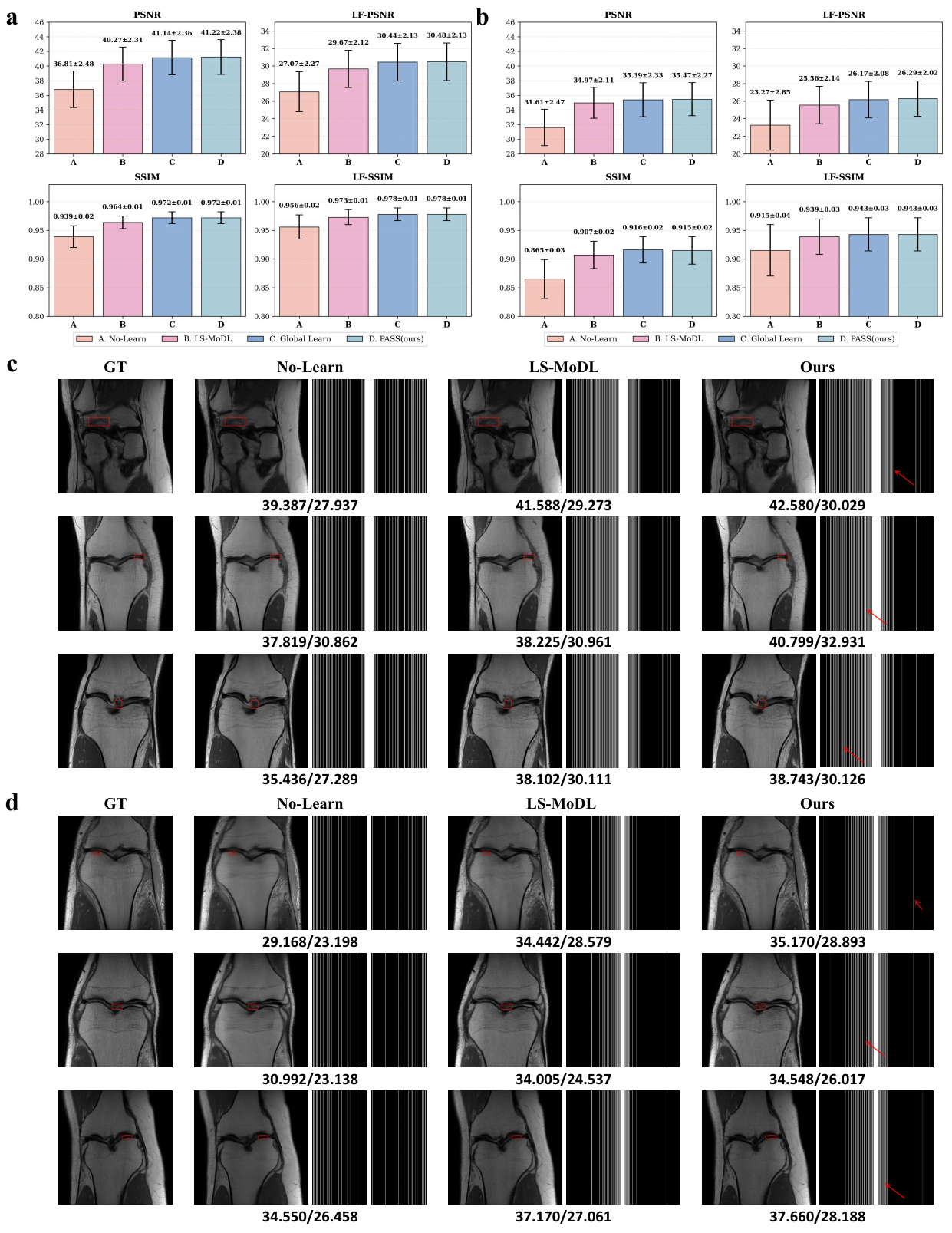}
	\caption{{Comparative analysis of sampling strategies for PD Knee images. Four strategies are evaluated, "No-learn": a fixed non-learnable mask, "LS-MoDL", a model-based baseline that jointly optimizes a reconstruction network and a learnable sampling mask, "Global Learn", a simplified variant of our framework employing LOUPE-based global sampling shared across all subjects without anomaly-aware sampling, and "Ours", the full PASS model integrating adaptive anomaly-aware sampling.
    a–b, Quantitative reconstruction results at $4\times$ and $8\times$ acceleration, respectively, including PSNR, SSIM, LF-PSNR, and LF-SSIM.
c–d, Visual comparisons of reconstructed images and corresponding sampling masks at $4\times$ and $8\times$ acceleration. Red arrows indicate regions where PASS adaptively increases sampling density in lesion-relevant k-space areas compared with other strategies.}}
\label{extend-fig4} 
\end{figure}

\begin{figure}[t]
	\centering
	\includegraphics[width=\textwidth]{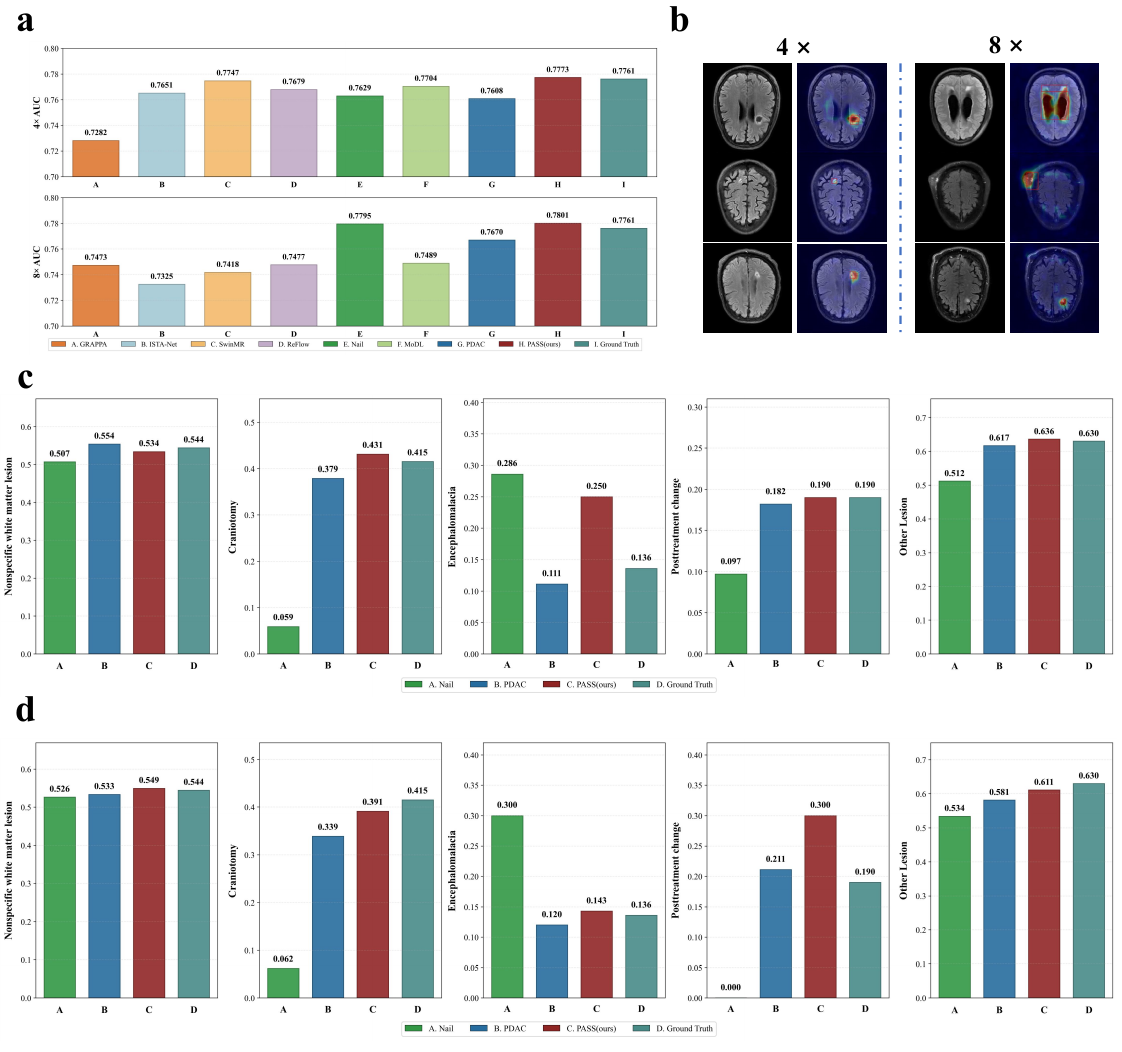}
	\caption{{Evaluation of downstream pathological tasks for FLAIR brain images. This figure compares the performance of different reconstruction methods on lesion-specific downstream tasks. a, Lesion classification AUC of a VLM model on reconstructed images generated by different methods under $4\times$ and $8\times$ acceleration, evaluated across all lesions collectively.
b, Localization of pathological regions by the VLM model on PASS-reconstructed images.
c–d, Multi-label classification accuracy for individual lesion types using a fine-tuned CLIP image encoder with an MLP classifier, at $4\times$ (c) and $8\times$ (d) acceleration, respectively.}}
\label{extend-fig5} 
\end{figure}

\begin{figure}[t]
	\centering
	\includegraphics[width=\textwidth]{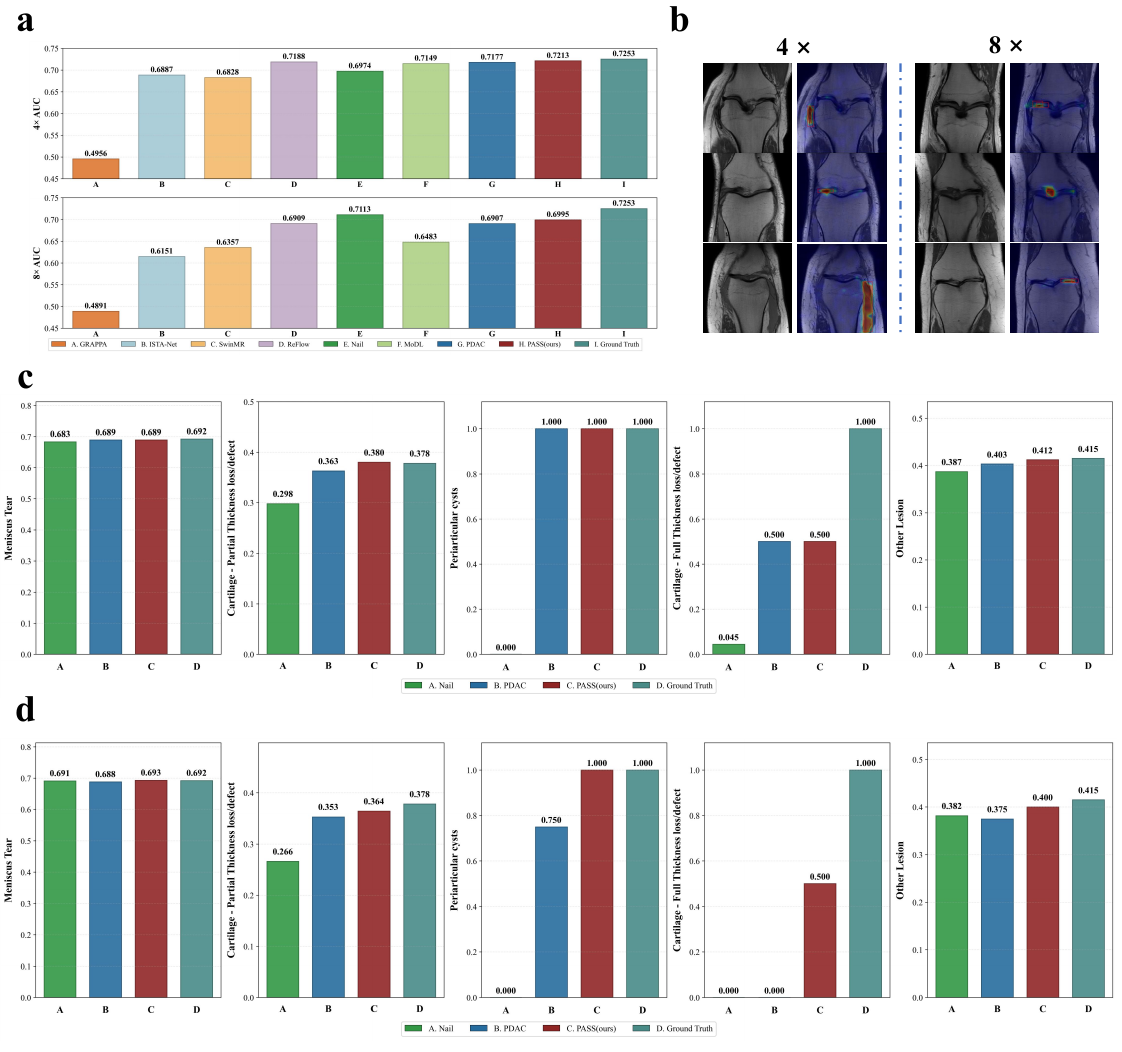}
	\caption{{Evaluation of downstream pathological tasks for PD Knee images. This figure compares the performance of different reconstruction methods on lesion-specific downstream tasks. a, Lesion classification AUC of a VLM model on reconstructed images generated by different methods under $4\times$ and $8\times$ acceleration, evaluated across all lesions collectively.
b, Localization of pathological regions by the VLM model on PASS-reconstructed images.
c–d, Multi-label classification accuracy for individual lesion types using a fine-tuned CLIP image encoder with an MLP classifier, at $4\times$ (c) and $8\times$ (d) acceleration, respectively.}}
\label{extend-fig6} 
\end{figure}